\documentclass[lettersize,journal]{IEEEtran}
\usepackage{amsmath,amsfonts}
\usepackage{algorithmic}
\usepackage{algorithm}
\usepackage{array}
\usepackage[caption=false,font=normalsize,labelfont=sf,textfont=sf]{subfig}
\usepackage{textcomp}
\usepackage{stfloats}
\usepackage{url}
\usepackage{verbatim}
\usepackage{graphicx}
\usepackage{cite}
\usepackage{subcaption}
\usepackage{subfig}
\usepackage{graphicx}
\usepackage{float}
\usepackage{cite}
\usepackage{makecell}
\usepackage{booktabs}   
\begin{document}

\title{P-MSDiff: Parallel Multi-Scale Diffusion for Remote Sensing Image Segmentation}

\author{Qi Zhang, Guohua Geng, Longquan Yan, Pengbo Zhou, Zhaodi Li, Kang Li and Qinglin Liu
\thanks{This research was funded by the the National Natural Science Foundation of China (62302393 and 62271393), Key Laboratory Project of Ministry of Culture and Tourism of China (1222000812). Key Research and Development Program of Shaanxi Province (2019GY215, 2021ZDLSF06-04). (Corresponding authors: Guohua Geng).}
\thanks{Qi Zhang, Guohua Geng, Longquan Yan, Kang Li and Qinglin Liu are with the School of Information Science and Technology, Northwest University, Xi’an, China (email:yin.shang1995@gmail.com; ghgeng@nwu.edu.cn; 18829512640@163.com; likang@nwu.edu.cn; lql@stumail.nwu.edu.cn).}
\thanks{Pengbo Zhou is with the School of Art and Media, Beijing Normal University, Beijing, China (email: zhoupengbo@bnu.edu.cn).}
\thanks{Zhaodi Li is with the School of Management, Nanjing Police University, Nanjing, China (email: 826976112@qq.com).}}

\markboth{Journal of \LaTeX\ Class Files,~Vol.~14, No.~8, August~2021}%
{Shell \MakeLowercase{\textit{et al.}}: A Sample Article Using IEEEtran.cls for IEEE Journals}


\maketitle

\begin{abstract}
Diffusion models and multi-scale features are essential components in semantic segmentation tasks that deal with remote-sensing images. They contribute to improved segmentation boundaries and offer significant contextual information. U-net-like architectures are frequently employed in diffusion models for segmentation tasks. These architectural designs include dense skip connections that may pose challenges for interpreting intermediate features. Consequently, they might not efficiently convey semantic information throughout various layers of the encoder-decoder architecture. To address these challenges, we propose a new model for semantic segmentation known as the diffusion model with parallel multi-scale branches. This model consists of Parallel Multiscale Diffusion modules (P-MSDiff) and a Cross-Bridge Linear Attention mechanism (CBLA). P-MSDiff enhances the understanding of semantic information across multiple levels of granularity and detects repetitive distribution data through the integration of recursive denoising branches. It further facilitates the amalgamation of data by connecting relevant branches to the primary framework to enable concurrent denoising. Furthermore, within the interconnected transformer architecture, the LA module has been substituted with the CBLA module. This module integrates a semidefinite matrix linked to the query into the dot product computation of keys and values. This integration enables the adaptation of queries within the LA framework. This adjustment enhances the structure for multi-head attention computation, leading to enhanced network performance and CBLA is a plug-and-play module. Our model demonstrates superior performance based on the J1 metric on both the UAVid and Vaihingen Building datasets, showing improvements of 1.60\% and 1.40\% over strong baseline models, respectively. Code and models are publicly available at: https://github.com/QiZhangsama/P-MSDiff
\end{abstract}

\begin{IEEEkeywords}
Diffusion model, multi-scale feature extraction, linear attention, remote sensing
\end{IEEEkeywords}

\section{Introduction}
\IEEEPARstart{R}{emote} sensing image segmentation is a critical research area within remote sensing image processing. Its primary objective is to divide remote sensing imagery into separate semantic regions. The utilization of high-resolution or hyperspectral images for precise delineation provides essential spatial information for various applications such as geological exploration, environmental monitoring, and urban planning. The utilization and implementation of high-resolution remote sensing data are becoming more demanding in terms of accuracy and efficiency due to the progress in satellite and drone technologies. In response, researchers have devised a variety of deep learning-based semantic segmentation algorithms tailored for remote sensing imagery. The effective integration and swift advancement of deep convolutional neural networks (CNNs)\cite{RN1, RN2, RN3} have resulted in remarkable performance across a range of tasks, such as image classification\cite{RN4,RN5}, image segmentation\cite{RN6,RN7}, instance segmentation\cite{RN8,RN09}, image vectorization\cite{RN10}, and object detection\cite{RN11,RN12,RN13}.

In recent years, advancements in semantic segmentation algorithms have followed three main trajectories. Fully Convolutional Networks (FCNs)\cite{RN14} and U-net\cite{RN15} leverage Convolutional Neural Networks (CNNs) for feature extraction in their encoding architectures. These models map interconnected features to semantic information using symmetric or asymmetric decoding structures to achieve accurate pixel-wise localization. On the other hand, frameworks based on transformers\cite{RN18,RN50,RN51} break down images into tokens and utilize self-attention\cite{RN25,RN26} mechanisms to dynamically assign different weights to individual elements within the input tokens. This approach enhances the ability to capture long-range dependencies among pixels. In semantic segmentation tasks, the input sequence usually comprises pixels contained in an image, while the output corresponds to the semantic category assigned to each pixel. This is exemplified by models such as GroupViT\cite{RN41}, SETR\cite{RN28}, TransUNet\cite{RN53}, and SegFormer\cite{RN29}. Diffusion models\cite{RN32,RN49} have demonstrated notable success in image segmentation by iteratively disseminating semantic information throughout images via a diffusion process. This approach aids in capturing contextual relationships among pixels. Throughout this process, the semantic correlation among pixels is taken into account, which enhances the efficiency of pixel-level classification.

To address semantic segmentation tasks across various data types, diffusion models commonly integrate masks into the diffusion process using conditional generative frameworks. Dmitry Baranchuk \textit{et al}.\cite{RN39} have suggested a method for extracting semantic information from input images by analyzing the intermediate activation layers of the network during the reverse diffusion process. This approach involves obtaining pixel representations through training on a restricted dataset. However, this inverse process is specifically designed for the original input images, leading to the omission of mask information during the diffusion restoration phase. Wu \textit{et al}.\cite{RN40} utilized dynamic conditional encoding to determine state-adaptive conditions for individual sampling steps. They also incorporated a Feature Frequency Parser to reduce the influence of high-frequency noise components in the sampling procedure. Kolbeinsson \textit{et al}.\cite{RN30} proposed an end-to-end semantic segmentation diffusion model that improves information capture by incorporating recursive denoising and hierarchical multi-scale enhancement processes. The UNetformer\cite{RN16} framework incorporates multiple linear attention\cite{RN31} modules. While these modules effectively decrease computational complexity in comparison to the dot-product attention\cite{RN54} mechanism, they demonstrate a uniform attention distribution that limits the ability to concentrate on intricate spatial features. Furthermore, the Linear Attention matrix exhibits a relatively low rank, making its linear dot-product computations highly susceptible to input noise, thereby limiting the diversity of features.

The static U-net-like architecture commonly found in diffusion networks restricts the receptive field, hindering the comprehensive gathering of global information. While the U-Net architecture incorporates skip connections to enhance information flow, it is limited in capturing long-range dependencies, which may result in reduced performance for tasks that involve processing extensive dependencies. Furthermore, the encoder-decoder architecture of U-net models depends on larger image dimensions for object detection. Kolbeinsson\cite{RN30} suggested mitigating model sensitivity to variations in features by incorporating multi-scale inputs; however, the backbone architecture does not include a specific strategy for processing multi-scale inputs.

Addressing the aforementioned concerns, this study presents a parallel multi-scale diffusion model that employs WNetFormer as its foundation. This model replaces the traditional single branch with parallel structures operating at different scales. This approach facilitates hierarchical multi-scale learning and the integration of additional information from branches at each diffusion stage. The methodology is elaborated in Section 3, and subsequent sections include comparative and ablation experiments carried out on the UAVid\cite{RN23} and Vaihingen Buildings\cite{RN24} datasets.

The principal contributions of this paper can be summarized as follows:

1) A parallel multi-scale diffusion model with interconnected branches is proposed for denoising, which enhances dimension-specific feature integration.

2) A cross-bridge linear attention mechanism is proposed to dynamically improve the weight distribution of the multi-head attention mechanism.

3) Our model demonstrated state-of-the-art precision metrics on both the UAVid and Vaihingen datasets. Ablation studies have been conducted to validate the effectiveness of our modules.

The subsequent sections of this paper are structured as follows: Section 2 offers a concise review of the current research in remote sensing image segmentation. Section 3 delineates our proposed parallel multi-scale diffusion model and elucidates the fundamental components of our methodology. In Section 4, we provide a detailed analysis of comparative and ablation experiments conducted on our method, along with the presentation of qualitative and quantitative results. Section 5 concludes the study by providing a summary of the findings and presenting suggestions for future research directions in this particular field.

\section{Related Work}
\subsection{Remote Sensing Image Semantic Segmentation}

In recent years, there have been notable advancements in the field of remote sensing imagery due to the progress of semantic segmentation algorithms that are based on deep learning models. Since the inception of Convolutional Neural Network (CNN) architectures for image segmentation applications, researchers have been consistently enhancing the efficiency and effectiveness of models customized for particular downstream tasks and data formats. This includes efforts to tackle end-to-end automatic segmentation challenges through FCNs and U-nets. Furthermore, in order to augment the representational capabilities of networks, dilated convolutions\cite{RN47, RN48} have been utilized to expand the receptive field of individual convolutional operations, thereby incorporating a greater amount of contextual features. The multi-layer structure of dilated convolutions may introduce gridding effects\cite{RN57}. To address this issue, \cite{RN55, RN56} proposed skip connections to incorporate low-level features into corresponding scale decoder networks. This approach effectively mitigates the loss of detail and spatial resolution resulting from downsampling operations. Moreover, it enhances the model's capacity to capture terrestrial boundaries and fine structures. In order to reduce the model's susceptibility to features with different scales, PSPNet\cite{RN21} and DeepLab\cite{RN22} have incorporated pyramid pooling modules to encompass a wide array of global information. GAN\cite{RN34} networks have been implemented to improve the effectiveness and versatility of segmentation models by utilizing adversarial learning. They incorporate conditional generative mechanisms to ensure that both generators and discriminators make optimal use of input data throughout the training and inference processes. This approach leads to semantic segmentation results that closely mimic real labels. Recent research has investigated various self-attention mechanisms in models such as Vision Transformer (VIT)\cite{RN18}, DETR\cite{RN19}, and TransUnet\cite{RN20}. These mechanisms aim to capture the relevance of categories and positions among image pixel tokens, enabling the learning of long-distance dependencies for segmentation tasks to uphold global semantic coherence.

In contrast to natural image segmentation tasks, remote sensing imagery poses notable challenges because of its high resolution and intricate land cover categories. These challenges are further compounded by the extensive incorporation of both distant and proximate information. Consequently, given the significant scale variations and stylistic diversity observed across various geographical regions, UAVFormer\cite{RN17} has introduced an approach that utilizes a multi-head self-attention transformer with aggregating windows. This method aims to dynamically integrate contextual and local information to alleviate semantic ambiguities. UNetFormer\cite{RN16} employs a global-local attention mechanism to capture information within the decoder, thereby improving the delineation of semantic segmentation boundaries in remote sensing images. Moreover, RNDiff\cite{RN30} introduced a hierarchical multi-scale recursive diffusion model to tackle significant scale variations, leading to cutting-edge outcomes.

\subsection{Diffusion Model}

Denoising Diffusion Probabilistic Models (DDPMs)\cite{RN32} are generative models that have the ability to generate high-quality samples. Recent studies have shown the superior performance of these models compared to other types of generative models\cite{RN33}. This has resulted in their extensive utilization in current research spanning various domains including image generation\cite{RN37}, image editing\cite{RN38}, text analysis\cite{RN58}, and speech recognition\cite{RN59}.

\begin{figure*}[!t]
	\centering
	\includegraphics[width=6.5in]{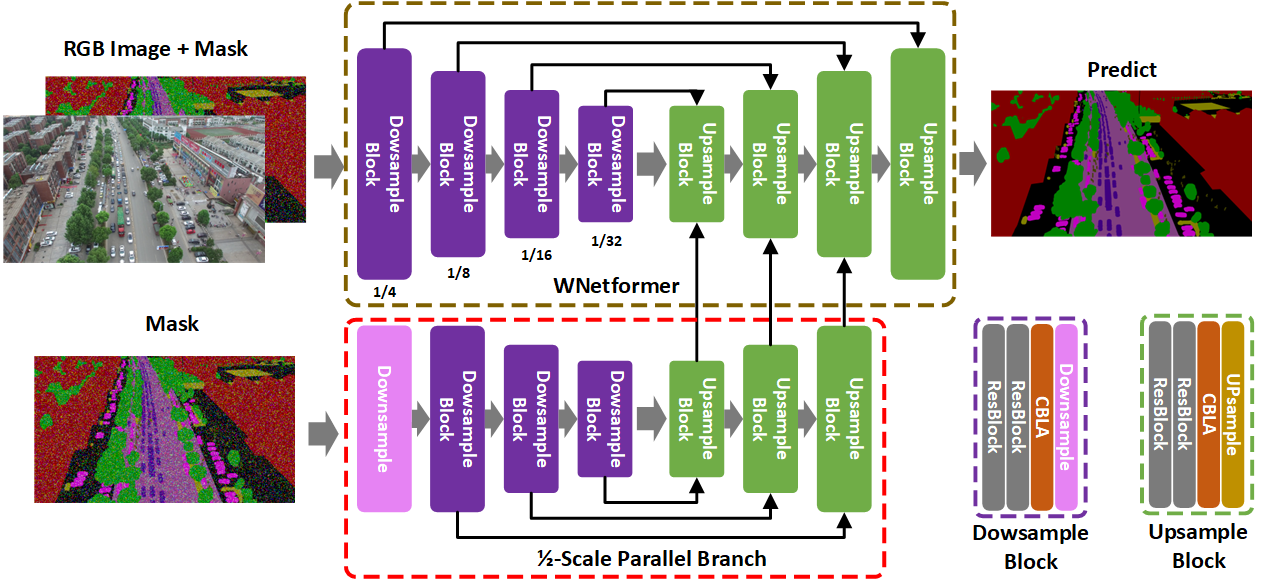}
	\caption{The overall framework of the P-MSDiff network. The network architecture comprises the WNetFormer core in the upper section and smaller parallel computational branches in the lower section. Feature fusion at consistent scales is denoted by black arrows. The Cross-Bridge Linear Attention (CBLA) mechanism is integrated into each module within the intermediary encoding and decoding structures to improve self-attention.}
	\label{fig_1}
\end{figure*}

Diffusion models are primarily comprised of two stages: a forward noising process and a reverse denoising process. In the forward process $q({x}_1:T|{x}_0)$, Gaussian noise is gradually introduced to the image using a Markov chain determined by $\beta_1, \ldots, \beta_T$, thereby transforming the initial data into an isotropic Gaussian distribution through a fixed-step $T$ noise diffusion mechanism. The process of gradual forward diffusion can be defined as follows:

\begin{equation}
	q(\mathbf{x}_{1:T}\mid\mathbf{x}_0)=\prod_{t=1}^Tq(\mathbf{x}_t\mid\mathbf{x}_{t-1})
\end{equation}

\begin{equation}
q(\mathbf{x}_t\mid\mathbf{x}_{t-1})=\mathcal{N}\left(\mathbf{x}_t;\sqrt{1-\beta_t\mathbf{x}_{t-1}},\beta_t\mathbf{I}\right)
\end{equation}

Wherein $\beta_t \in (0,1) $ represents the variance, $x_1, \ldots, x_T$ are latent variables, and $T$ denotes the time steps.

During the denoising procedure of reverse diffusion, the elimination of noise is progressively acquired by the Markov chain of the joint distribution $p_\theta(x_{0:T})$, expressed as:

\begin{equation}
p_{\theta}(\mathbf{x}_{0:T})=p_{\theta}(\mathbf{x}_{T})\prod_{t=1}^{T}p_{\theta}(\mathbf{x}_{t-1}\mid\mathbf{x}_{t})
\end{equation}

\begin{equation}
p_\theta(\mathbf{x}_{t-1}\mid\mathbf{x}_t)=\mathcal{N}\left(\mathbf{x}_{t-1};\mu_\theta(\mathbf{x}_t,t),\mathbf{\Sigma}_\theta(\mathbf{x}_t,t)\right)
\end{equation}

Classical diffusion theory is commonly utilized for high-quality generative tasks. However, the significant increase in computational power consumption has prompted researchers to predominantly leverage its pre-trained weights for downstream tasks that share similar data types. This preference substantially limits the utilization of diffusion models in the context of remote sensing image prediction. RNDiff\cite{RN30} incorporates spatial and semantic features extracted from CNNs as conditional embeddings in the diffusion process. This approach pioneers an end-to-end semantic segmentation model for diffusion, which is trained from the beginning, thereby aiding in remote sensing image segmentation by reducing noise. Our study extends the RNDiff by using it as a baseline and improving denoising performance through a stacked branching approach. Further elaboration on this methodology will be provided in Section 3.

\subsection{Multi-Branch Structure}

Multi-branch architectures represent a prevalent design paradigm in neural networks, distinguished from single-path networks by their additional branches designed to capture differential features from the main network. These features include, but are not limited to, modality\cite{RN42}, depth\cite{RN43}, context\cite{RN44}, edges\cite{RN45}, texture\cite{RN46}, and abstract representations. For instance, in change detection tasks, SNUNet-CD\cite{RN36} employs a dual-path branch of a Siamese neural network to process satellite land cover images from different years, conducting differential analysis at the terminal stage to derive change prediction probabilities. Feng et al.\cite{RN35} introduces a multi-branch classification fusion structure with an attention mechanism, adaptively assigning weights based on the discriminative capabilities of different branches. Overall, multi-branch structures are instrumental in extracting and utilizing feature diversity, offering broad applicability in handling multimodal information, hierarchical features, and surrounding contexts.

Multi-branch architectures are a prominent design paradigm in neural networks, characterized by the presence of multiple branches that are intended to capture distinct features from the primary network. These features encompass various aspects such as modality\cite{RN42}, depth\cite{RN43}, context\cite{RN44}, edges\cite{RN45}, texture\cite{RN46}, and abstract representations. For example, in change detection tasks, SNUNet-CD\cite{RN36} utilizes a dual-path branch of a Siamese neural network to analyze satellite land cover images from different years. It performs a differential analysis at the final stage to generate change prediction probabilities. Feng et al.\cite{RN35} study presents a multi-branch CNN including attention moudle for hyperspectral image classification that incorporates an adaptive region search. The network features a multi-branch classification fusion structure with an attention mechanism that dynamically assigns weights according to the discriminative abilities of individual branches. Multi-branch structures play a crucial role in extracting and leveraging feature diversity, providing extensive utility in processing multi-modal information, hierarchical features, and contextual surroundings.

\subsection{Linear Attention (LA)}

The self-attention mechanism has shown remarkable versatility and effectiveness in various deep learning models, serving as a fundamental component of the transformer\cite{RN60} architecture. The performance is improved by incorporating additional weight parameters, enabling the model to dynamically learn attention weights for individual elements within the input sequence. This process efficiently captures the interrelations among the elements of the sequence.

In self-attention mechanisms, the memory and computational costs associated with dot product operations exhibit a quadratic increase in relation to the input size. This phenomenon significantly limits the feasibility of training high-resolution remote sensing images. Shen et al.\cite{RN31} proposed a linear attention mechanism that accomplishes automatic normalization of $QK^T$ through accumulated softmax operations. Its form of attention is:

\begin{equation}
LA(Q,K,V)=softmax_2(Q)softmax_1(K)^TV
\end{equation}

The terms ${softmax}_1$ and ${softmax}_2$ denote the softmax operations conducted along the first and second dimensions, respectively.

\begin{figure*}[!t]
	\centering
	\includegraphics[width=6.5in]{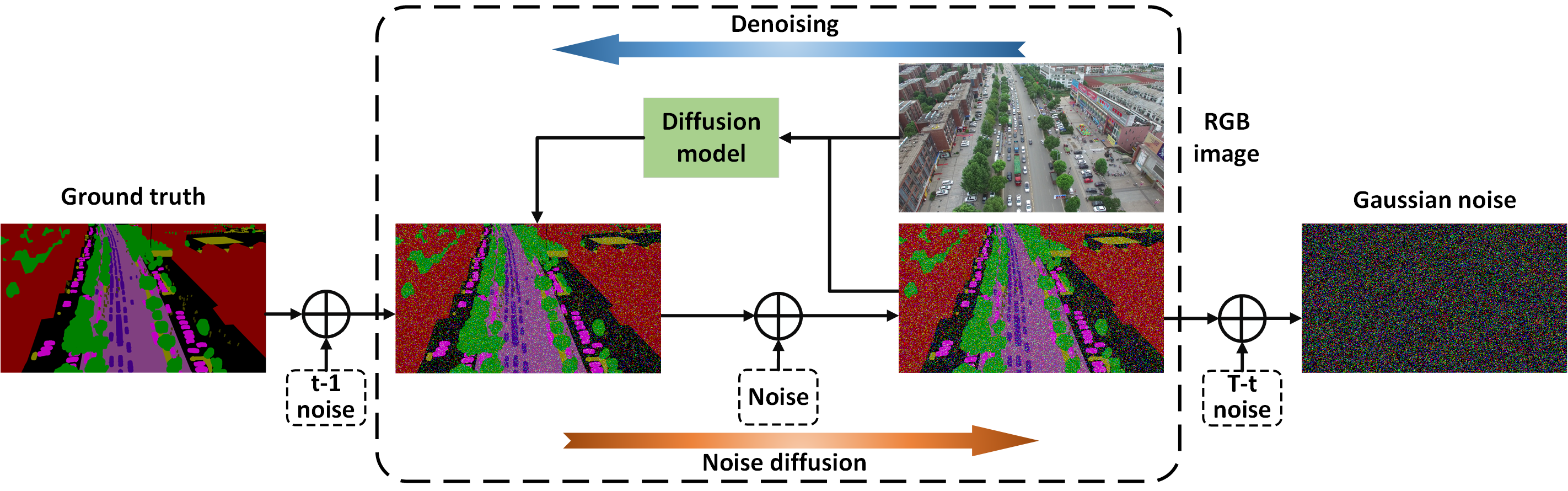}
	\caption{Training process of remote sensing image diffusion models. The central structure applies $T$ steps of noise diffusion to the ground truth, consequently utilizing the diffusion model for $T$  steps of denoising training.}
	\label{fig_2}
\end{figure*}

\section{Methods}

In this section, we first present the comprehensive framework suggested in this research. This is followed by a detailed explanation of the proposed Parallel Multi-Scale Diffusion (P-MSDiff) structure and the plug-and-play Cross-Bridge Linear Attention (CBLA) module.
 
\subsection{Overview}
The network model proposed, P-MSDiffNet, is illustrated in Figure \ref{fig_1}. The framework is built upon the RNDiff model, with the incorporation of parallel multi-scale computational branches integrated into its backbone structure, WNetFormer. Subsequently, the linear self-attention mechanism in the transformer architecture is replaced with the plug-and-play Cross-Bridge Linear Attention (CBLA) module. Ultimately, a composite loss function is utilized to train the proposed framework.

\subsection{Background Of Semantic Segmentation Diffusion}

To support semantic segmentation tasks for remote sensing imagery, we introduce the input image $Img$ and its associated segmentation mask $M$, denoted at time step $t\in[0,T]$. Therefore, the gaussian noise introduced at each time step $t$ can be expressed as:

\begin{equation}
N_t=\mathcal{N}(0,\beta_t\mathbf{I})
\end{equation}

The mask produced by the forward diffusion at time step $t$ can be characterized as:

\begin{equation}
q(M_t\mid M_{t-1})=M_{t-1}+N(0,\beta_t\mathbf{I})
\end{equation}

The noise schedule  $\beta_t$ is linearly correlated with $t$.

\begin{equation}
\beta_t=\delta(t)
\end{equation}

During the reverse process, for each denoising step, the model $D$ is trained as follows:

\begin{equation}
D(t-1|t,Img,M_t)\approx M_t-M_{t-1}
\end{equation}

This, in turn, results in a recursive process:

\begin{equation}
M_0\approx M_T-D(T-1\mid T,Img,M_t)-\cdots-D(0\mid1,Img,M_1)
\end{equation}

Wherein, the $M_0$ state denotes the segmentation result, while the $M_T$ state corresponds to pure gaussian noise.

The particular procedures for forward and reverse training are illustrated in Figure \ref{fig_2}. Given an original image, along with associated labels and a designated time step denoted as $T$, a consistent noising protocol is implemented. This protocol involves perturbing the input labels gradually over $T$ steps until they reach a state of pure noise. Throughout each stage of the diffusion process, the diffusion model acquires image representations to assess the noise level of the mask, consequently recovering the labels to generate segmentation predictions.

\subsection{Parallel Multi-Scale Diffusion (P-MSDiff)}

Specifically, a variant of UnetFormer, named WNetFormer, is utilized as the foundational segmentation model, replicating the structure outlined in Section 2. This network is equipped with dual inputs, enabling the simultaneous processing of RGB images and noise masks. WNetFormer integrates a hybrid model of residual convolution and linear self-attention mechanisms to capture contextual and detailed information within objects of varying resolution sizes.

WNetFormer employs ResNet50 as its encoder, establishing a network with a "W" shape by integrating a four-stage encoding design with a fully symmetrical decoding structure. This configuration exhibits similarities with UNet-like networks by consolidating category and semantic information in the deep encoding feature space and gradually recovering size and intricate details in the decoding phase. During the diffusion stage, the image embeddings and mask embeddings are integrated to create a dual input for the network.

During the training phase, RNDiff integrates RGB images at three distinct scales as inputs, and then utilizes the diffusion model to refine the initial outputs for accurate category prediction. Nevertheless, during the processing of the coarse outputs, only single-scale contextual information is incorporated, mainly originating from the noise simulation phase of the model. In the context of remote sensing imagery encompassing vast areas and intricate spatial arrangements, the utilization of multi-scale noise learning has the capability to identify nuanced characteristics and contextual hints during the restoration procedure. This approach consequently improves the efficacy and resilience of the model in managing scenarios involving multiple categories and varying scales.

Convolutional Neural Networks (CNNs) demonstrate a preference for processing noise in isolation rather than concurrently managing noise and semantic information. The filter structures within each layer of convolution can autonomously execute denoising functions, whereas the depiction of land cover categories' distribution relies solely on deep features. Consequently, structures have been devised to specifically handle noise processing at various scales. As depicted in Figure \ref{fig_1}, a WNetFormer network at a 1/2 scale was utilized as a parallel branch, serving as a substructure of the WNetFormer network. In contrast to the baseline configuration, when presented with a coarse output characterized by C and N denoting the number of channels and categories, and H and W representing height and width, the initial step involves linearly reducing it to a scale of (1/8)H×(1/8)W. This reduction results in a feature size for the input that is half the size of the original network. This is succeeded by three symmetrical sets of downsampling and upsampling layers. During the process of upsampling, the regenerated features at scales of 1/32, 1/16, and 1/8 are successively integrated with the corresponding dimensions of the primary backbone via concatenation and dimension reduction. This parallel denoising branch has the capability to directly handle small-sized input masks without necessitating any feature transformation. This feature enables the independent learning of low-resolution noise patterns.

Ultimately, the decoding segments of the branches are concatenated with those of WNetFormer, followed by the diffusion process, which can be described as follows:

\begin{equation}
	C_i=[W_i:B_i]
\end{equation}

Wherein $W_i$ denotes the $i-th$  decoding stage of WNetFormer and $B_i$ represents the $i-th$  decoding stage of the branch, with $i\in[1:3]$ in the 1/2 scale branch structure.

By utilizing a model with a consistent architecture as the branch, the model's fitting capability and generalization performance are minimally impacted by the multi-branch structure during the recursive denoising training process. Likewise, a parallel 1/4 scale branch was constructed; nevertheless, the excessively small scale of information processed resulted in the lack of positive effects in the feature fusion process. Comprehensive findings and analyses will be delineated in Section 4.5.

The improved multi-branch structure preserves an end-to-end network model configuration, enabling it to follow the original diffusion model's training protocol consistently throughout the training process. This adaptable adjustment can be considered a universal enhancement strategy for diffusion networks, rendering it widely applicable to analogous tasks and associated models.

\begin{figure*}[!t]
	\centering
	\subfloat[Linear Attention]{\includegraphics[width=3.3in]{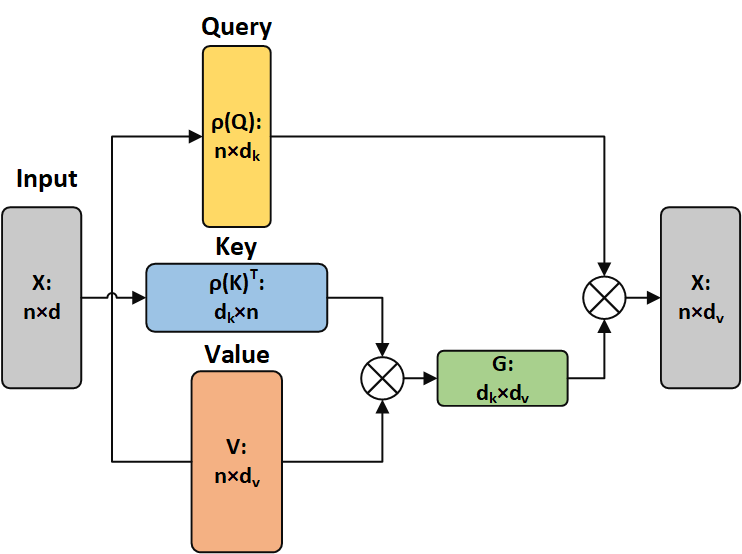}%
}
	\hfil
	\subfloat[Cross-Bridge Linear Attention]{\includegraphics[width=3.3in]{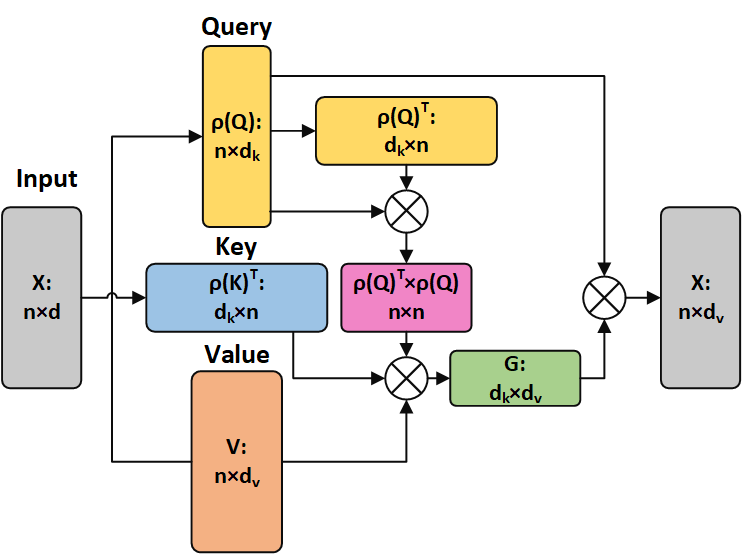}%
}
	\caption{A comparison of the LA module and the CBLA module. Each box symbolizes an input, output, or intermediate matrix. The symbol $\rho$ is utilized to indicate normalization through the softmax function. $N$, $d$, $d_k$, and $d_v$ denote the magnitude of the input and the dimensions of the input, keys, and values, respectively.
 }
	\label{fig_3}
\end{figure*}

\subsection{Cross-Bridge Linear Attention(CBLA)}

The linear attention\cite{RN31} mechanism accomplishes automatic normalization by employing cumulative softmax operations. It has been theoretically demonstrated to be equally efficient as dot-product attention. The attention formulation is presented as follows:

\begin{equation}
LA(Q,K,V)=softmax_2(Q)softmax_1(K)^TV
\end{equation}

The configuration of Linear Attention (LA) is illustrated in Figure \ref{fig_3} (a). By linearly aggregating keys (K) along with values (V), LA avoids dot-product aggregation method with queries (Q), thereby accomplishing dimensionality reduction within module $G$. Nevertheless, the transformation from $QK^T$ to $K^TV$ overlooks the incorporation of adaptive attention towards query features. While queries are ultimately aggregated with $G$ based on weights, the dimensionality reduction in $G$ may not guarantee the preservation of all global contextual information. Certain essential features may be compromised as a result of dimensionality reduction, features that are pivotal for comprehending global information. This situation may result in specific challenges in the model's management of worldwide connections.

This paper addresses a significant gap in linear attention (LA) by introducing a cross-bridge linear attention mechanism that incorporates query self-attention. This mechanism enhances aggregation computations by creating an intermediate self-attention matrix for $Q$ to interact with the $K$ and $V$ matrices. The paper introduces the intermediate value $Q'$ to measure the self-similarity of queries, referred to as the self-attention bridge of $Q$, which is mathematically expressed as:

\begin{equation}
Q'=softmax_2(Q)^Tsoftmax_2(Q)
\end{equation}

As illustrated in Figure \ref{fig_3} (b), the CBLA attention mechanism involves the transformation of input feature vectors through three linear layers to generate $query (Q)$, $key (K)$, and $value (V)$ representations. Subsequently, a positive semi-definite matrix $Q'$ is generated by performing the dot product operation between $Q$ and its transpose, which captures the self-attention expressions of $Q$. During the process of linear aggregation of $K$ and values $V$, the dot product operation is initially conducted between $K$ and $Q'$, and subsequently with $V$. As the dimension of $Q'$ is $n×n$, it does not affect the final matrix calculation dimension in the computation of $KTQ'V$. The computational process can be articulated as follows:

\begin{equation}
CBLA(Q,K,V)=softmax_2(Q)softmax_1(K)^TQ^{\prime}V
\end{equation}

For the input feature map $X$, the module flattens it into a matrix $X\in\mathbb{R}^{n_{h\times w}\times d}$, applies the CBLA attention mechanism, and reshapes the result into $h\times w\times d_v$. 

\subsection{Loss Function}

The diffusion network model employed for iterative noise restoration facilitates end-to-end training of the process, starting from Gaussian noise to the predicted outcomes. To simultaneously address noise restoration and measure segmentation standards, mean squared error (MSE) and cross-entropy functions are commonly utilized as the loss functions for the diffusion model.

\begin{equation}
\mathcal{L}_{diff}=\mathcal{L}_{mse}+\mathcal{L}_{ce}
\end{equation}

The specific forms of $\mathcal{L}_{mse}$ and $\mathcal{L}_{ce}$ are as follows:

\begin{equation}
\mathcal{L}_{mse}=\dfrac{1}{N}\sum_{i=1}^N\left(n_i^{gt}-n_i^p\right.)^2
\end{equation}

\begin{equation}
\mathcal{L}_{ce}=-\frac{1}{N}\sum_{i=1}^{N}\sum_{c=1}^{C}y_{i,c}\log(\hat{y}_{i,c})
\end{equation}

Wherein, $n^{gt}$ denotes the standard-added noise, $n^p$ represents the noise prediction, $N$ signifies the total number of pixels in the image, and $C$ denotes the preset classes in the segmentation.

Nevertheless, the allocation of land cover categories in remote sensing images frequently demonstrates class imbalance, resulting in a long-tail effect. To address segmentation errors resulting from imbalanced class distribution, we utilize the Dice loss as a supplementary loss function in formulating a composite loss. The formula for the Dice loss is as follows:

\begin{equation}
	\mathcal{L}_{dice}=1-\frac{1}{C}\sum_{c=1}^{C}\frac{2\times\sum_{i}^{N}\hat{y}_{i,c}\times y_{i,c}}{\sum_{i}^{N}\hat{y}_{i,c}^{2}+\sum_{i}^{N}y_{i,c}^{2}}
\end{equation}

Hence, the composite loss can be expressed as:

\begin{equation}
\mathcal{L}_{diff}=\mathcal{L}_{mse}+\lambda\mathcal{L}_{ce}+(1-\lambda)\mathcal{L}_{dice}
\end{equation}

In this study, the experimental set value for $\lambda$ is 0.2.

\section{Experiments}

To assess the effectiveness of our proposed methodology, comprehensive experiments were carried out using the publicly accessible UAVid and Vaihingen segmentation datasets. In this section, an overview of the datasets and experimental setups is presented. Subsequently, comparative experiments were conducted on both datasets, along with a comprehensive presentation of the qualitative outcomes obtained from different comparative methodologies. Subsequently, the datasets were subjected to comprehensive ablation experiments.

\subsection{Experimental setting} 
\subsubsection{DATASETS}
Comparative and ablation experiments were performed on two publicly available datasets, namely the UAVid dataset and the Vaihingen dataset.
 
The UAVid dataset consists of 30 video sequences and is a high-resolution drone semantic segmentation dataset. The sequences encompass 420 high-resolution images captured at a 4K resolution from an oblique perspective. This dataset comprises 200 images for training, 70 for validation, and 150 for testing purposes. The dataset has been classified into eight semantic classes, namely buildings, roads, trees, low vegetation, moving cars, static cars, people, and others, to facilitate semantic segmentation tasks.

The Vaihingen dataset showcases urban scenes from an aerial viewpoint, encompassing a variety of detached and small multi-story buildings. The dataset comprises 33 remote sensing images of different dimensions, encompassing near-infrared, red, and green bands. For the experiments, the dataset was divided into 168 images of equal dimensions. The initial 100 images were allocated for training purposes, while the remaining 68 images were reserved for testing to assess the model's performance. In the process of categorization, we have designated the main buildings as the exclusive focus of recognition, while all other categories have been grouped together as a background class.

\subsubsection{Evaluation Metrics}

The Intersection over Union (IoU) metric, also referred to as the Jaccard Index, was implemented for semantic segmentation. This metric includes the calculation of the mean IoU (mIoU) across all categories.

\begin{equation}
	IoU=\frac{TP}{TP+FP+FN}
\end{equation}

\begin{equation}
	mIoU=\frac1N\sum_{i=1}^N\frac{TP_i}{TP_i+FP_i+FN}_i
\end{equation}

Simultaneously, we utilized the F1-score as an additional metric.

\begin{equation}
	F1=\frac{2TP}{2TP+FP+FN}
\end{equation}

In this context, $TP$ (True Positive) represents correctly identified positives, $FP$ (False Positive) indicates incorrectly identified positives, $FN$ (False Negative) denotes incorrectly identified negatives, and $N$ denotes the total number of classes within the dataset.

\subsubsection{Diffusion structure}

Diffusion structure plays a crucial role in determining the final segmentation outcomes, especially in the context of developing multi-branch networks. We have minimized structural deviation to align with established parameters through the creation of the WNetFormer network. This network integrates the CBLA module along with its half-scale submodules. The structure proposed in this study does not hinder the primary training process of the diffusion model. It offers a thorough demonstration of the effectiveness of the CBLA module and P-MSDiff when compared to the baseline. Moreover, quarter-scale submodules were implemented to streamline the development of varied multi-branch network structures, and the appropriateness of scaling branches was investigated through ablation studies.

\subsubsection{Training Details}

The training of diffusion models usually requires longer time steps for optimization. The empirical parameters defining the baseline have undergone experimental validation by RNdiff, which has resulted in the selection of a time step of 128. To augment the training dataset, we utilized random horizontal flips (50\%) and made adjustments to brightness, contrast, saturation, and hue. The models were implemented and trained utilizing the PyTorch framework on an AMD Ryzen 9 5950X with 64GB RAM and an NVIDIA GeForce RTX 3090. The training process spanned 100 epochs with a batch size of 4. The AdamW optimizer was employed with an initial learning rate of $10^{-5}$. It utilized a cosine annealing strategy for adjusting the learning rate, resetting to the initial rate after every 10 epochs, and implementing a weight decay of 0.01. High-definition input images were segmented with overlap. Subsequently, both the original images and labels were organized into batches of size 512×512 for network training. Our training experiments have demonstrated that the proposed model can achieve convergence and the expected segmentation results within 80 epochs.

\subsection{Binary segmentation} 

In this section, experiments are conducted on the Vaihingen Buildings dataset to evaluate the effectiveness of the proposed approach. Following the segmentation of the original remote sensing images, buildings located at the center of each batch are assigned as the target category, while the remaining areas are classified as background. This approach effectively simulates a binary segmentation problem. Despite the simplistic nature of this categorical approach, the intricate scenes within the dataset present substantial challenges to our model. The results from both quantitative and qualitative analyses indicate that our approach attains superior performance in the binary segmentation task using the Vaihingen Buildings dataset.

\subsubsection{Comparative Methods}

In order to assess the efficacy of our methodology, we conduct a comparative analysis with five traditional semantic segmentation algorithms designed for remote sensing imagery in this experimental section.

\textbf{Deep Structured Active Contours (DSAC)}\cite{RN61} incorporates prior knowledge and constraints into the segmentation process using a CNN framework. It utilizes a polygon model that is based on constraints and priors, which is designed for end-to-end training.

\textbf{Trainable Deep Active Contours (TDAC)}\cite{RN62} establishes an automated image segmentation framework through the integration of Trainable Deep Active Contours with a CNN, enabling a seamless and adaptive segmentation procedure.

\textbf{DARNet}\cite{RN63}  utilizes a deep convolutional neural network as its foundational framework for the prediction of energy maps. These energy maps are then employed to derive energy functions for the purpose of image segmentation.

\textbf{SegDiff}\cite{RN64} presents the integration of diffusion models with U-Net for iterative semantic segmentation of remote sensing imagery, demonstrating the efficacy of diffusion-based approaches in this field.

\textbf{RNDiff}\cite{RN30} integrates the transformer architecture to establish the WNetFormer framework for recursive diffusion, yielding notable segmentation outcomes through its pioneering methodology.

\begin{figure*}[!t]
	\centering
	\includegraphics[width=7in]{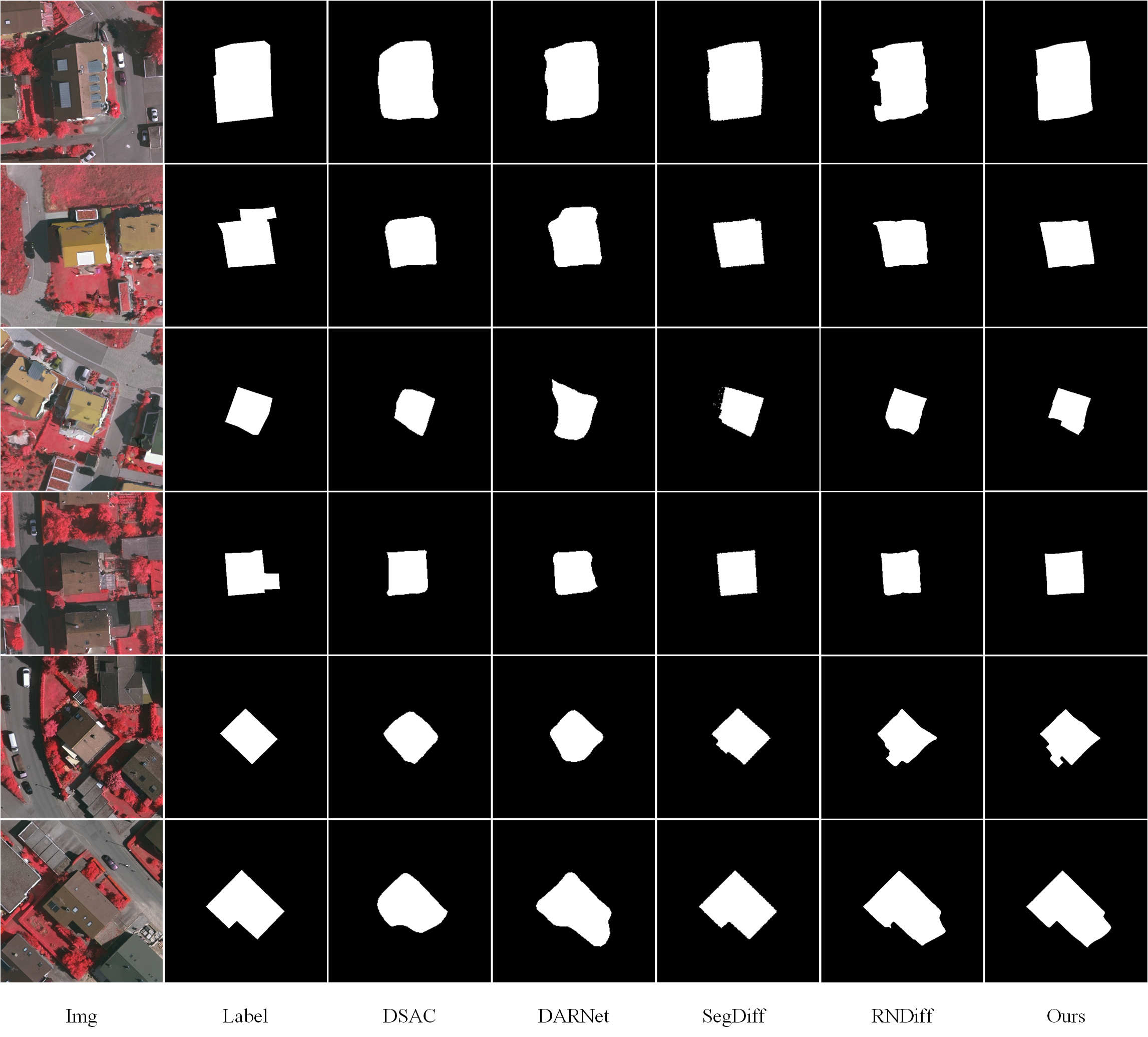}
	\caption{Visualization on the Vaihingen Buildings dataset. Our method exhibits more standard edges and achieves results closer to the ground truth labels in the recognition of main buildings, presenting a more precise visual performance compared to CNN networks and other diffusion models.}
	\label{fig_4}
\end{figure*}

Our model consists of several key components: The Time Step Encoding Module, which documents the correlation between diffusion times and model weights, and then incorporates this information into the encoding phase; the CNN-Transformer Encoder Architecture, which operates simultaneously during both the image encoding and segmentation mask encoding phases; the Decoding Process, which is closely integrated with the encoding phase, progressively upsamples while incorporating lower-dimensional features from the encoder of the same resolution; the Cross-Bridge Linear Attention (CBLA) Module, which is embedded in each downsampling stage of the encoder and each upsampling stage of the decoder; Sub-branches, with fewer structural layers, reconstruct the core of the encoder-decoder architecture, merging with the main framework at different points in the decoding process.

\subsubsection{Quantitative Results}

Table \ref{tab:table1} illustrates the binary segmentation performance on the Vaihingen buildings dataset. In contrast to tasks involving multiple categories, our primary focus lies in the segmentation of central buildings. We utilize mIoU and F1 metrics to measure the accuracy of the primary subject exclusively, highlighting the highest accuracies in bold.

A comparative analysis of the quantitative outcomes obtained from the methods mentioned above indicates that algorithms utilizing CNN structures, including DSAC, TDAC, and DARNet, demonstrate subpar performance in the identification of primary subjects. The limitations observed can be attributed to the inherent constraints of networks that rely solely on CNNs. These networks extract profound contextual information mainly through a series of convolution kernels, leading to a confined perceptual field and, consequently, a limited capacity to comprehend global contexts and capture long-range dependencies. When confronted with remote sensing imagery set against intricate backgrounds, convolutional neural networks frequently encounter challenges in distinguishing subtle features and semantic attributes within the images. Consequently, this difficulty results in lower mean Intersection over Union (mIoU) scores.

The SegDiff and RNDiff algorithms exhibit relatively superior performance. SegDiff utilizes the fitting and diffusion mechanisms of noise to effectively improve the CNN's capacity to detect and conceal intricate backgrounds. RNDiff utilizes transformers to enhance intricate spatial semantic correlations in high-dimensional data, thereby achieving superior segmentation performance through the establishment of supplementary global relationships.

In comparison to RNDiff, our approach demonstrated a 1.40\%enhancement in mIoU and a 0.25\% rise in F1-score when evaluated on the Vaihingen Buildings dataset. The marginal difference in the F1-score, which reached a final score of 98.93\%, is ascribed to the boundary effects that may emerge in binary segmentation tasks when optimization reaches such high levels.

\begin{table}[]
	\caption{Comparison of various methodologies for analyzing Vaihingen buildings. Results are presented as reported in the respective publication. F1-score was not reported in all original publications.} \label{tab:table1}
	\centering
	\begin{tabular}{c|cc}
		\toprule
		Method & mIoU & F1-score \\
		\midrule
		DSAC\cite{RN61} & 84.00 & - \\
		TDAC\cite{RN62} & 89.16 & - \\
		DARNet\cite{RN63} & 88.24 & - \\
		SegDiff\cite{RN64} & 91.12 & 95.14 \\
		RNDiff\cite{RN30} & 92.50 & 98.68 \\
		\textbf{Ours} & \textbf{93.90} & \textbf{98.93} \\
		\bottomrule
	\end{tabular}
\end{table}

\begin{figure*}[!t]
	\centering
	\includegraphics[width=7in]{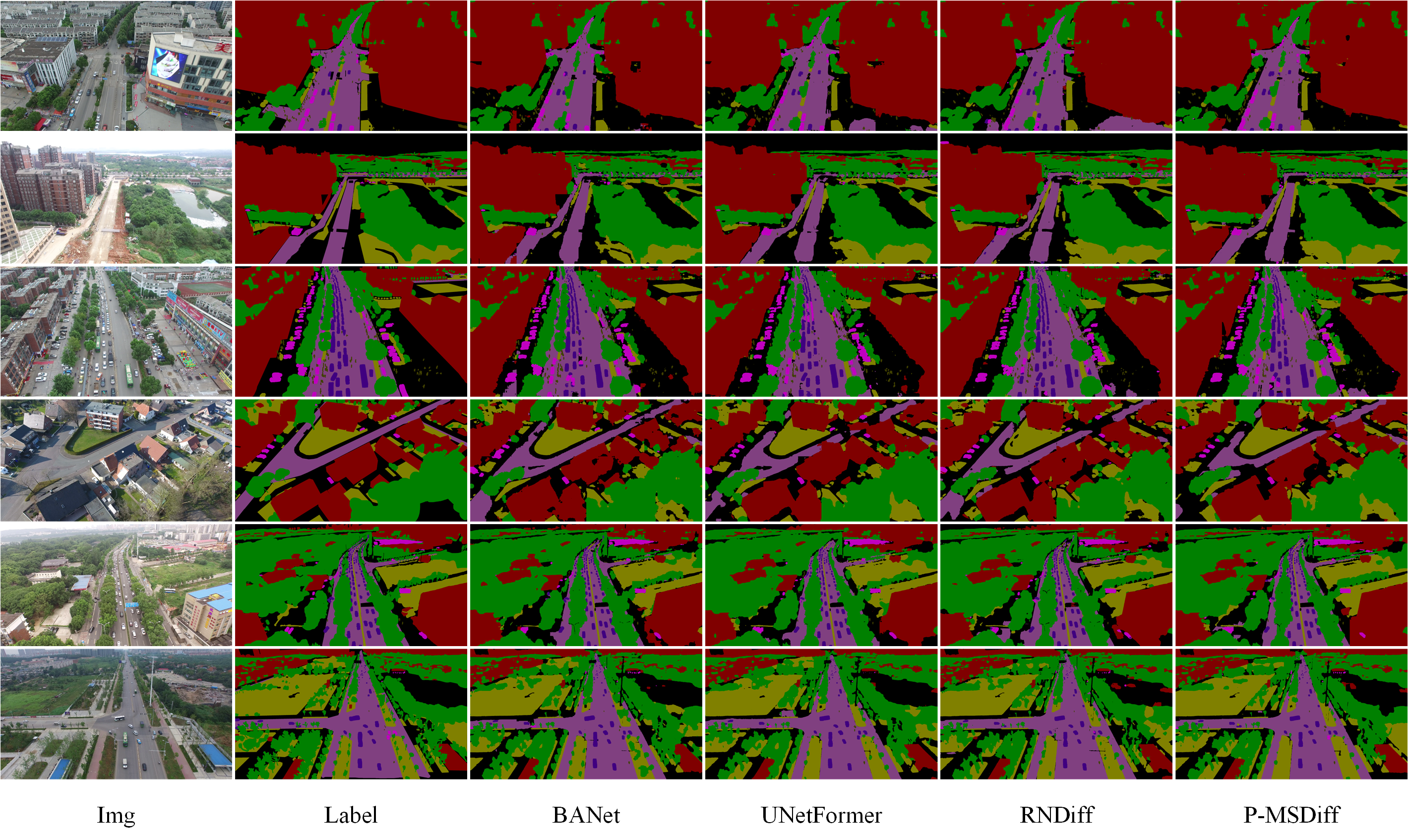}
	\caption{The segmentation results on the UAVid dataset. The experimental results of this approach are generally similar to RNDiff, showing consistent segmentation errors in large-scale classes. However, in the detection of small-scale objects such as Human, Moving Car and Static Car, it exhibits more precise semantic labeling.}
	\label{fig_5}
\end{figure*}

\begin{table*}[]
	\caption{A comparison of various methods on the UAVid test dataset. Each column represents the Intersection over Union (IoU) per respective class, with the rightmost column indicating the mean IoU.}\label{tab:table2}
	\centering
	\begin{tabular}{c|ccccccccc}
		\toprule
		Method & Building & Road & Tree & \makecell{Low \\ Vegetation} & \makecell{Moving \\ Car} & \makecell{Static \\ Car} & Human & Clutter & mIoU \\
		\midrule
		UAVFormer & 81.5 & 67.1 & 76.2 & 48.5 & 62.2 & 28.8 & 12.5 & 48.8 & 53.2 \\
		CANet & 86.6 & 62.1 & 79.3 & \textbf{78.1} & 47.8 & 68.3 & 19.9 & 66.0 & 63.5 \\
		BANet & 85.4 & 80.7 & 78.9 & 62.1 & 52.8 & \textbf{69.3} & 21.0 & 66.6 & 64.6 \\
		UNetFormer & 87.4 & \textbf{81.5} & \textbf{80.2} & 63.5 & 73.6 & 56.4 & 31.0 & \textbf{68.4} & 67.8 \\
		RNDiff & \textbf{87.7} & 80.1 & 79.8 & 63.5 & 71.2 & 60.1 & 26.3 & 68.3 & 67.1 \\
        \textbf{Ours} & 87.0 & 81.0 & 79.4 & 62.9 & \textbf{73.8} & 62.9 & \textbf{31.9} & 67.3 & \textbf{68.7}	\\					
		\bottomrule
	\end{tabular}
\end{table*}

\subsubsection{Quantitative Results}

Comparative segmentation results of the Vaihingen Buildings dataset are depicted in Figure \ref{fig_4}. The analysis indicates a reduced noise distribution and a strong alignment with the real-world scene. Segmentation techniques utilizing CNN, such as DSAC and DARNet, produce noise segmentation outcomes that are irregular and more fragmented, characterized by rounded corners. While these methods generally differentiate between the primary building structures and other background elements, they notably diverge from the authentic architectural forms. Instances of misidentification occur in the primary segmentation, where non-primary structures are erroneously identified as main components. The segmentation masks generated also exhibit irregular edges and a significant level of noise distribution.

Meanwhile, there has been an enhancement in the performance of diffusion model techniques, as evidenced by the capabilities of SegDiff and RNDiff in attenuating noise interference to a certain degree. Nevertheless, their effectiveness is constrained by various factors, such as the scale of information, leading to inadequate detail representation. This includes irregular serrations at building edges and misclassifications caused by factors like lighting conditions. The method we propose demonstrates superior performance compared to robust baselines and alternative networks. It effectively identifies and separates noise interferences originating from neighboring buildings, interconnected spaces, shadows, and lighting in the majority of the dataset samples. Moreover, it effectively delineates essential structural components that may not be explicitly labeled, showcasing the enhanced sensitivity of our network towards detecting small targets.

\subsection{Multi-class segmentation} 

In contrast to binary segmentation tasks, multi-class segmentation involves the pixel-level classification of two or more categories, posing greater challenges for the model due to the presence of inter-class correlations and mutual interference.

\subsubsection{Comparative Methods}

In the context of the multi-class segmentation task, we have incorporated four additional segmentation techniques for comparison.:

\textbf{UAVFormer}\cite{RN50} presents an adaptive integration of contextual and local information using a multi-head self-attention transformer with aggregation windows to address semantic ambiguities.

\textbf{CANet}\cite{RN49} establishes a dual-branch network comprising a high-resolution branch to effectively capture spatial details and a context branch incorporating a streamlined version of global aggregation and local distribution blocks to accurately capture contextual dependencies.

\textbf{BANet}\cite{RN44} introduces a bilateral awareness network that incorporates dependency and texture paths to effectively capture long-range connections and intricate details in high-resolution images.

\textbf{UNetFormer}\cite{RN45} utilizes a global-local attention mechanism to capture information in the decoder, aiming to improve the accuracy of semantic segmentation boundaries in remote sensing images.

\subsubsection{Quantitative Results}

Table \ref{tab:table2} presents the segmentation performance of P-MSDiff compared to other methods on the UAVid dataset, including mIoU and IoU for each class. The best performance is highlighted in bold.

Notably, our proposed P-MSDiff has achieved state-of-the-art results on the UAVid test dataset. The table reveals a recent focus in related research on the dependence of long-range contextual relationships, building comprehensive spatial information in large-scene classes. Both UnetFormer and RNDiff exhibit significantly high IoU performance in wide-range scene classes. However, segmentation results show poorer performance in tasks involving the recognition of small-scale classes, particularly those dominated by the Human class. This is mainly due to these classes' lower pixel proportions and strong individuality. While BANet has shown outstanding performance in the Static Car class, it must have corresponding accuracy in the relatively confusing Moving Car class.

Compared to the baseline method RNDiff, P-MSDiff demonstrates a more balanced performance across all classes. Although there is slightly decreased in IoU for large-scale classes, particularly in Building, Tree, and Low Vegetation, the decrease is minimal at 0.7\%, 0.4\%, and 0.6\%, respectively. Similar to its performance on the Vaihingen buildings dataset, P-MSDiff shows significant improvements in IoU for small targets such as Moving Car, Static Car, and Human on UAVid, with increases of 2.6\%, 2.8\%, and 5.6\% respectively. This improvement in IoU comes at a reasonable cost compared to the slight decrease observed in the baseline IoU, making it worthwhile. The final mIoU achieved is 68.7\%, representing a 1.40\% improvement over RNDiff.

\subsubsection{Qualitative Results}

\begin{figure*}[!t]
	\centering
	\includegraphics[width=7in]{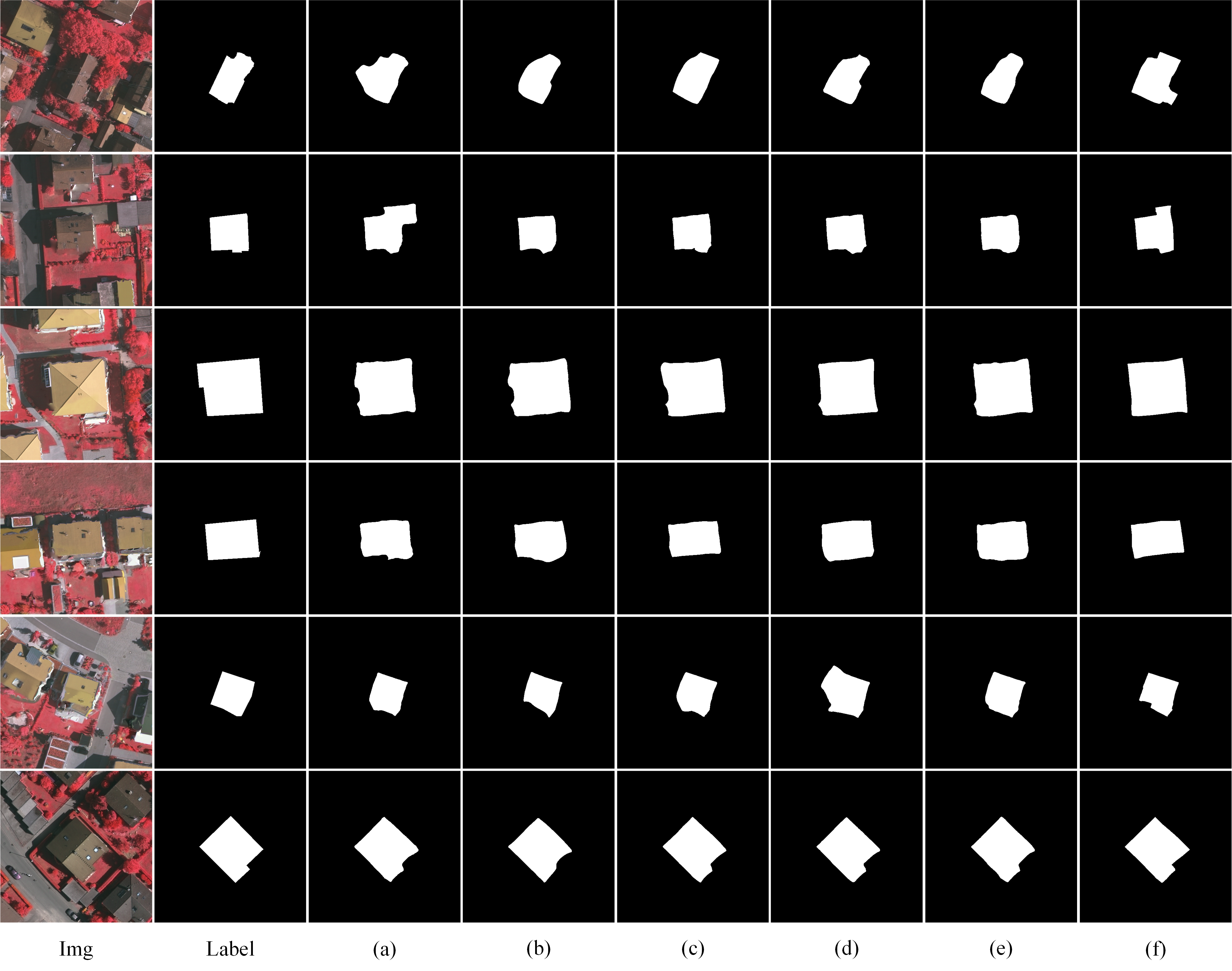}
	\caption{The ablation results conducted results on the Vaihingen revealed the impact of various modifications on the performance of the model. These modifications included: the following: (a) RNDiff; (b) Addition of 1/2 scale branch; (c) Addition of 1/2 scale and 1/4 scale branches; (d) Addition of 1/2 scale and 1/4 scale branches along with the CBLA module; (e) Implementation of the CBLA module; (f) Addition of 1/2 scale branch and the CBLA module.}
	\label{fig_6}
\end{figure*}

The visual segmentation results on the UAVid dataset, as depicted in Figure \ref{fig_5}, align with the quantitative findings. Our proposed method exhibits qualitatively similar segmentation errors to RNDiff in most large-scale categories, with a notable decrease in misidentification rate for the Road classes, avoiding confusion with Clutter scenes. Additionally, in segmenting similar classes like Moving Car and Static Car, P-MSDiff rectifies an error in RNDiff where a single Car may simultaneously include both moving and static states, ensuring each vehicle entity remains independent. This improvement is attributed to the finer-grained information the low-scale branch network captured. This refinement is evident in the segmentation boundaries of various category masks, where, apart from Tree, other classes show logically shaped and bounded visualizations with reduced noise points. Furthermore, from a holistic perspective, our proposed method exhibits smoother transitions at class junctions.

\begin{table*}[]
	\caption{Different settings on Vaihingen dataset.}\label{tab:table3}
	\centering
	\begin{tabular}{cccc|cc}
		\toprule
		Backbone & 1/2-Scale Branch & 1/4-Scale Branch & CBLA & mIoU & F1-score\\
		\midrule
		\checkmark & - & - & - & 92.50 & 98.68 \\
		\checkmark & \checkmark & - & - & 92.94 & 98.77 \\
		\checkmark & \checkmark & \checkmark & - & 92.79 & 98.73 \\
		\checkmark & \checkmark & \checkmark & \checkmark & 93.14 & 98.79 \\
		\checkmark & - & - & \checkmark & 93.34 & 98.83 \\
		\textbf{\checkmark} & \textbf{\checkmark} & \textbf{-} & \textbf{\checkmark} & \textbf{93.90} & \textbf{98.93} \\
		\bottomrule
	\end{tabular}
\end{table*}

\subsection{Ablation Studies} 

In this section, ablation studies were conducted on the text structure, with a primary focus on the parallel multi-scale branches and the CBLA module as key points of reference.

\subsubsection{Ablation Design}

Upon examination of the proposed P-MSDiff algorithm, we attribute its main benefits to the incorporation of the 1/2 scale sub-branch structure and the plug-and-play CBLA module. Table \ref{tab:table3} presents the mean Intersection over Union (mIoU) and F1-score metrics for different structural configurations.

In the context of the binary segmentation task using the Vaihingen dataset, we developed various versions of P-MSDiff structures for conducting ablation studies.

(a)	RNDiff: The original network remains unaltered.

(b)	P-MSDiff with an added 1/2 scale branch: As depicted in Figure \ref{fig_1}, the structure involves the concurrent operation of a sub-branch processing at a 1/2 scale, thereby expanding the fundamental framework of RNDiff.

(c)	P-MSDiff with 1/2 scale and 1/4 scale branches: This variation extends the approach described in (b) by incorporating an extra parallel sub-branch at a reduced scale.

(d)	P-MSDiff enhanced with 1/2 scale and 1/4 scale branches and the CBLA module: This version replaces the LA module in (c) with the CBLA module.

(e)	RNDiff with integrated CBLA module: This setup involves substituting the LA module in RNDiff with the CBLA module.

(f)	P-MSDiff with 1/2 scale branch and CBLA Module: This version replaces the LA module in (b) with the CBLA module.

The examination of sub-branch scales indicates that the 1/4 scale sub-branch, responsible for handling very low-resolution mask noise, might not always have a beneficial impact on the segmentation results in the ultimate feature reconstruction.

\subsubsection{Result Analysis}

A comparison was conducted on the qualitative and quantitative results among various configurations. The outcomes obtained from parallel branches and the CBLA module exhibit a higher level of semantic accuracy in segmentation when contrasted with networks that have a single branch. The statement suggests that the incorporation of parallel branch structures and modules enhances the ability to capture and interpret intricate semantic information present in images. In this architectural framework, parallel branches are dedicated to distinct levels of features and semantic nuances. The CBLA module plays a crucial role in refining and adjusting the representations of local areas by leveraging information from these branches. This process enhances the accuracy of the final model output in delineating subject boundaries and internal structures within images. This configuration offers notable benefits in addressing boundary ambiguity and objects with comparable semantics but varying structures, thereby improving the precision and resilience of segmentation. Furthermore, in qualitative experiments, these configurations demonstrated exceptional performance in terms of mean Intersection over Union (mIoU) and F1-score metrics.

Secondly, the qualitative results of incorporating a 1/4 scale sub-branch suggest that the integration of an extra low-resolution channel brings forth semantic details of secondary elements, such as shadows, vegetation, and terrain associated with buildings. This aspect of performance deterioration is also evident in the quantitative findings. In the optimal network configuration of P-MSDiff with a 1/2 scale branch and the CBLA module, the delineation of building boundaries yields the most favorable outcomes, and the segmentation results closely align with the ground truth labels. While certain segmentation outcomes may contain category errors, the mIoU performance remains state-of-the-art.

\section{Conclusion}

This paper introduces a P-MSDiff network as a solution for semantic segmentation in remote sensing images. The P-MSDiff network effectively captures multi-scale information from input images of various resolutions by employing a multi-branch architecture. By stacking multiple branches within the P-MSDiff module, the network denoises features of images at various resolutions concurrently. This process guarantees the extraction of semantic information and shallow textures during segmentation. The parallel multi-scale information integrates a variety of features, offering a more robust basis for semantic segmentation tasks. Moreover, the CBLA module reconstructs the computational framework of linear multi-head attention by dynamically resetting the adaptive weight allocation of queries within the transformer framework. It dynamically adapts the weights of segmentation masks to account for varying scale information, thereby substantially improving the resilience of the segmentation model. Extensive experiments conducted on the UAVid dataset and the Vaihingen building dataset illustrate the efficacy of the proposed P-MSDiff network in comparison to other contemporary methodologies.

The increasing prevalence of diffusion models in remote sensing land cover analysis is anticipated to expand with additional research. The utilization of high-resolution and hyperspectral imagery poses increased challenges for model implementation. Through continuous research and innovation, it is anticipated that diffusion models will enhance their ability to accommodate diverse remote sensing data, thereby providing more precise and comprehensive geographic information support for the sustainable development of industrial deployment, geological exploration, and satellite detection sectors.

\bibliographystyle{IEEEtran}
\bibliography{references}

\vfill

\end{document}